\documentclass[10pt,natbib,floatsintext,jou]{apa6}
\usepackage[utf8]{inputenc}
\usepackage{amsmath}
\usepackage{graphicx}
\usepackage{authblk}
\usepackage{hyperref}

\makeatletter
\providecommand{\@LN}[2]{}
\makeatother
\title{Rapid trial-and-error learning with simulation supports flexible tool use and physical reasoning}
\leftheader{Allen, Smith, Tenenbaum}
\shorttitle{Rapid trial-and-error learning in physical reasoning}

\author[1,2,*]{Kelsey R. Allen}
\author[1,2,*]{Kevin A. Smith} 
\author[1,2]{Joshua B. Tenenbaum}

\affil[1]{Department of Brain and Cognitive Sciences, MIT}
\affil[2]{Center for Brains, Minds and Machines (CBMM)}
\affil[*]{Indicates equal contribution}
\affiliation{~} 

\authornote{\vspace{-1em}\\
Corresponding author:\\
Kelsey R. Allen\\
Department of Brain and Cognitive Sciences\\
Massachusetts Institute of Technology\\
77 Massachusetts Avenue, Cambridge, MA 02139\\
krallen@mit.edu}

\abstract{
Many animals, and an increasing number of artificial agents, display sophisticated capabilities to perceive and manipulate objects. 
But human beings remain distinctive in their capacity for flexible, creative tool use -- using objects in new ways to act on the world, achieve a goal, or solve a problem.
To study this type of general physical problem solving, 
we introduce the Virtual Tools game.
In this game, people solve a large range of challenging physical puzzles in just a handful of attempts.
We propose that the flexibility of human physical problem solving rests on an ability to imagine the effects of hypothesized actions, while the efficiency of human search arises from rich action priors which are updated via observations of the world.
We instantiate these components in the ``Sample, Simulate, Update'' (SSUP) model and show that it captures human performance across 30 levels of the Virtual Tools game. 
More broadly, this model provides a mechanism for explaining how people condense general physical knowledge into actionable, task-specific plans to achieve flexible and efficient physical problem-solving.
}


\hypersetup{
    colorlinks=true,
    linkcolor=blue,
    filecolor=magenta,      
    urlcolor=blue,
    citecolor=black,
}
\begin{document}

\maketitle
While trying to set up a tent on a camping trip, you realize that the ground is too hard for the tent stakes, and you have no hammer. What would you do? You might look around for a suitable hammer substitute, passing over objects like pinecones or water bottles in favor of a graspable rock. And if that rock failed to drive in the stakes at first, you might try a different grip, or search for a heavier rock. Most likely, you would only need a handful of attempts before you found an approach that works.

Determining how to pound in tent stakes without a hammer is an example of the flexibility and efficiency of more general physical problem solving.
It requires a causal understanding of how the physics of the world works, and sophisticated abilities for inference and learning to construct plans that solve a novel problem. 
Consider how, when faced with the tent stake challenge, we do not choose an object at random; we choose a rock because we believe we know how we could use it to generate sufficient force on the stake.
And if we find that the first rock fails, we again search around for a solution, but use the knowledge of our failures to guide our future search. This style of problem solving is a very structured sort of trial-and-error learning: our search has elements of randomness, but within a plausible solution space, such that the goal can often be reached very quickly.

\begin{figure}[ht!]
    \centering
    \includegraphics[width=0.95\linewidth]{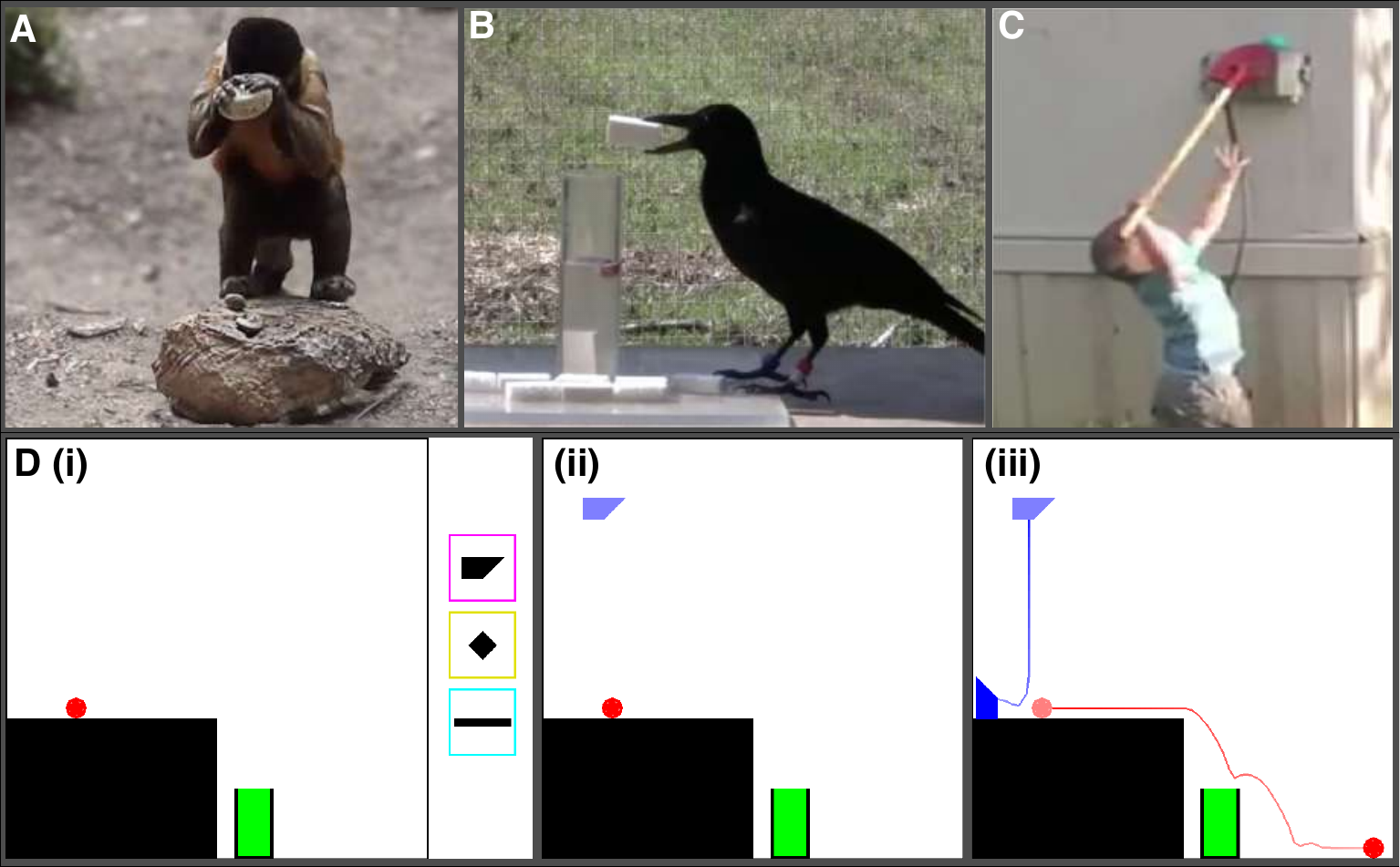}
    \vspace{0.1em}
    \caption{
    Examples of using objects to achieve a goal.
    \textbf{(A)} Bearded capuchin monkey opening a cashew nut with an appropriately sized stone \cite[photo by Tiago Falótico;][]{luncz2016wild}.
    \textbf{(B)} New Caledonian crow using heavy blocks to raise the water level in a tube in order to retrieve food \cite[][]{jelbert2014using}.
    \textbf{(C)} Toddler using a shovel to reach a ball (from  \href{https://youtu.be/hwrNQ93-568?t=198}{youtu.be/hwrNQ93-568?t=198}).
    \textbf{(D)} One illustrative trial in the Virtual Tools game (\href{https://sites.google.com/view/virtualtoolsgame}{https://sites.google.com/view/virtualtoolsgame}). (i) The player must get the red object into the green goal using one of the three tools. (ii) The player chooses a tool and where to place it. (iii) Physics is turned ``on'' and the tool interacts with other objects. The action results in a near miss.
    }
    \vspace{-1em}
    \label{fig:exp_example}
\end{figure}

Here we study the cognitive and computational underpinnings of flexible tool use. 
While human tool use relies on a number of cognitive systems -- for instance, knowing how to grasp and manipulate an object, or understanding how a particular tool is typically used -- here we focus on ``mechanical reasoning,'' or the ability to spontaneously repurpose objects in our environment to accomplish a novel goal \cite[][]{osiurak2016tool,orban2014neural,goldenberg2009neural}.

We target this mechanical reasoning because it is the type of tool use that is quintessentially human. 
While other animals can manipulate objects to achieve their aims, only a few species of birds and primates have been observed to spontaneously use objects in novel ways, and we often view these activities as some of the most ``human-like'' forms of animal cognition \cite[e.g., Fig.~\ref{fig:exp_example}A,B;][]{shumaker2011animal}.
Similarly, while AI systems have become increasingly adept at perceiving and manipulating objects, none perform the sort of rapid mechanical reasoning that people do.
Some artificial agents learn to use tools from expert demonstrations \cite[][]{xie2019improvisation}, which limits their flexibility. Others learn from thousands of years of simulated experience \cite[][]{baker2019emergent}, which is significantly longer than required for people. Still others can reason about mechanical functions of arbitrary objects but require perfect physical knowledge of the environment \cite[][]{toussaint2018differentiable}, which is unavailable in real-world scenarios.
In contrast, even young humans are capable tool users: by the age of four they can quickly choose an appropriate object and determine how to use it to solve a novel task \cite[e.g., picking a hooked rather than straight pipe cleaner to retreive an object from a narrow tube, Fig.~\ref{fig:exp_example}C;][]{beck2011making}.

\begin{figure*}[t!]
    \centering
    \includegraphics[width=0.9\textwidth]{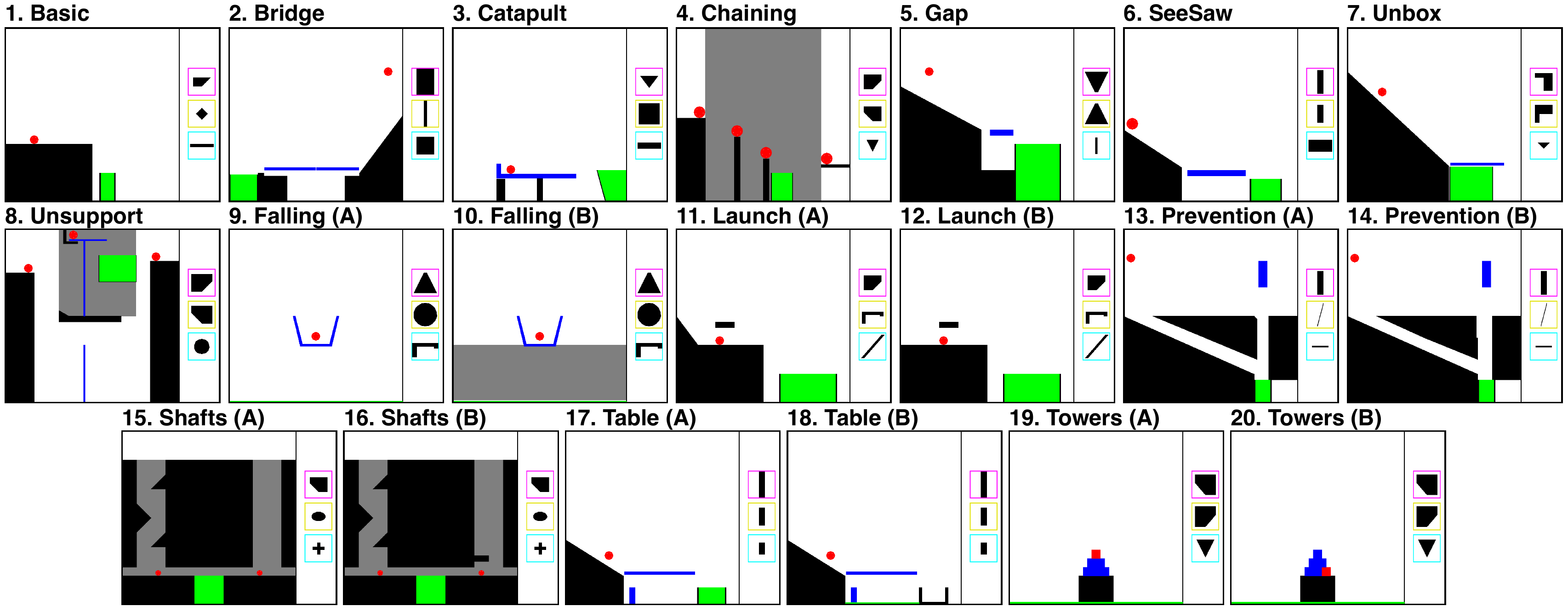}
    \caption{Twenty levels used in the Virtual Tools game. Players choose one of three tools (shown to the right of each level) to place in the scene in order to get a red object into the green goal area. Black objects are fixed, while blue objects also move; grey regions are prohibited for tool placement. Levels denoted with A/B labels are matched pairs.}
    \vspace{-1em}
    \label{fig:trials}
\end{figure*}

What are the cognitive systems that let us use tools so flexibly, and accomplish our goals so rapidly?
It has been suggested that mechanical reasoning relies on mental simulation, which lets us predict how our actions will cause changes in the world \cite[][]{osiurak2016tool}.
This general purpose simulation is a necessary component that supports our ability to reason about objects in novel environments, but by itself cannot explain how we make and update our plans so quickly.
We propose that another key to rapid tool use is knowing what sorts of actions to even consider -- both from an initial understanding of what actions are useful, and by updating this belief from observing the outcome of our actions, in simulation and in reality.

This paper makes two contributions. First, we introduce the Virtual Tools game, which presents a suite of physical problem solving challenges, and allows for precise, quantifiable comparisons between human and machine agents. Second, we present a minimal model of flexible tool use, called ``Sample, Simulate, Update'' (SSUP). This model is built around an efficient albeit noisy simulation engine that allows the model to act \emph{flexibly} across a wide variety of physical tasks. 
To solve problems \emph{rapidly}, the SSUP model contains rich knowledge about the world in the form of a structured prior on candidate tools and actions likely to solve the problem, which allows it to limit its simulations to promising candidates.
It further learns from its simulations and from observing the outcome of its own actions to update its beliefs about what those promising candidates should be.
Across 30 Virtual Tools levels in two experiments, we show that an instantiation of the SSUP model captures the relative difficulties of different levels for human players, the particular actions performed to attempt to solve each level, and how the solution rates for each level evolve. 

\subsection*{The Virtual Tools game}
Inspired by human tool use, as well as mobile physics games \cite[][]{brainiton}, we propose the Virtual Tools game as a platform for investigating the priors, representations, and planning and learning algorithms used in physical problem solving  (\href{https://sites.google.com/view/virtualtoolsgame}{https://sites.google.com/view/virtualtoolsgame}). 
This game asks players to place one of several objects (``tools'') into a two-dimensional dynamic physical environment in order to achieve a goal: getting a red object into a green region (Fig.~\ref{fig:exp_example}D). This goal is the same for every level, but what is required to achieve it varies greatly. 
Once a single tool is placed, the physics of the world is enabled so that players see the effect of the action they took.
If the goal is not achieved, players can ``reset'' the world to its original state and try again; they are limited to a single action on each attempt. 
We designed 30 levels -- 20 for the original experiment (Fig.~\ref{fig:trials}) and 10 for a validation experiment (Fig.~\ref{fig:cv_fig}A) -- to test concepts such as `launching', `blocking', and `supporting'.
Of the first 20 levels, 12 were constructed in six `matched pairs' which incorporated small differences in the goals or objects in the scene to test whether subtle differences in stimuli would lead to observable differences in behavior. 

\begin{figure*}[ht!]
    \centering
    \includegraphics[width=0.95\textwidth]{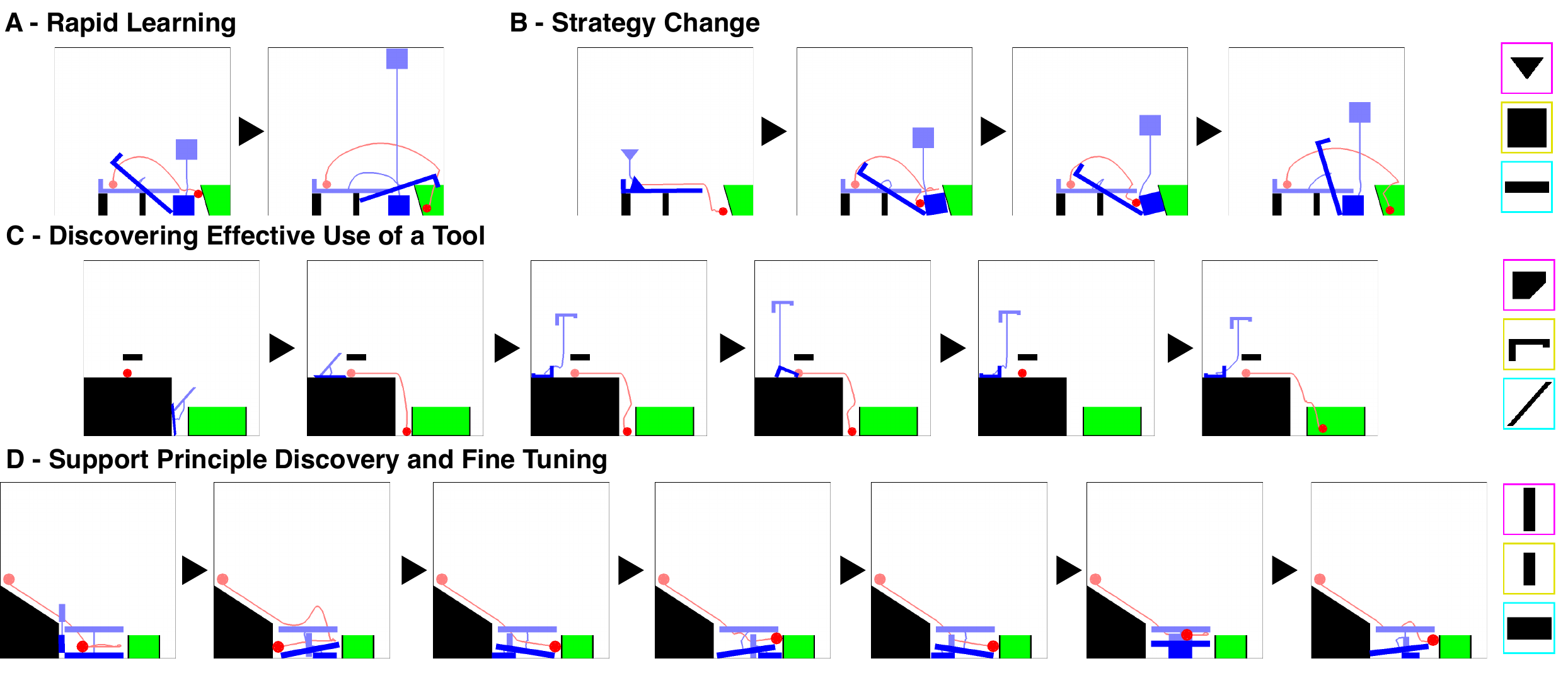}
    \caption{Examples of participants' behavior on three levels, representative of rapid trial-and-error learning: Initial plans are structured around objects, followed by exploring to identify more promising strategies and then refining actions until success. Objects start as shown by light blue/red outlines and 
    follow paths traced out by colored lines. Possible tool choices shown to the right. 
    \textbf{(A)} In the Catapult level, a useful strategy is often identified immediately and rapidly fine-tuned. 
    \textbf{(B)} Other participants first try an unsuccessful strategy but then switch to a more viable strategy and refine it. 
    \textbf{(C)} The Launch (B) level is designed to prevent obvious solutions. 
    This participant may have initially believed the ball would start rolling and attempted to use a tool as a bridge.
    When this failed they realized they needed to launch the ball, but only discovered after several trials how to use a tool in a non-obvious way to accomplish this, via a hooking motion around the blocking ledge. They then took several more trials to fine-tune this action. 
    \textbf{(D)} In the SeeSaw level, a participant realized on the second attempt they must support the platform for the ball to roll across, then tried different ways of making this happen.}
    \label{fig:slowfast}
\end{figure*}

The Virtual Tools game presents particular challenges that we believe underlie the kinds of reasoning required for rapid physical problem solving more generally.
First, there is a \textbf{diversity of tasks} that require different strategies and physical concepts to solve, but employ shared physical dynamics that approximate the real world.
Second, the game requires \textbf{long-horizon causal reasoning}. 
Since players can only interact with the game by placing a single object, they must be able to reason about the complex cause and effect relationships of their action long into the future when they can no longer intervene.
Finally, the game elicits \textbf{rapid trial-and-error learning} in humans. 
Human players do not generally solve levels on their first attempt, but also generally do not require more than 5-10 attempts in order to succeed. 
People demonstrate a wide range of problem-solving behaviors, including ``a-ha'' insights where they suddenly discover the right idea for how to solve a particular task, as well as incremental trial-and-error strategy refinement.
Figure~\ref{fig:slowfast} demonstrates how this occurs in practice, showing four different examples of participants learning rapidly or slowly, and discovering different ways to use the tools across a variety of levels.

\subsection*{\textit{Sample, Simulate, Update} Model (SSUP)}
\begin{figure*}[t!]
    \centering
    \includegraphics[width=0.8\textwidth]{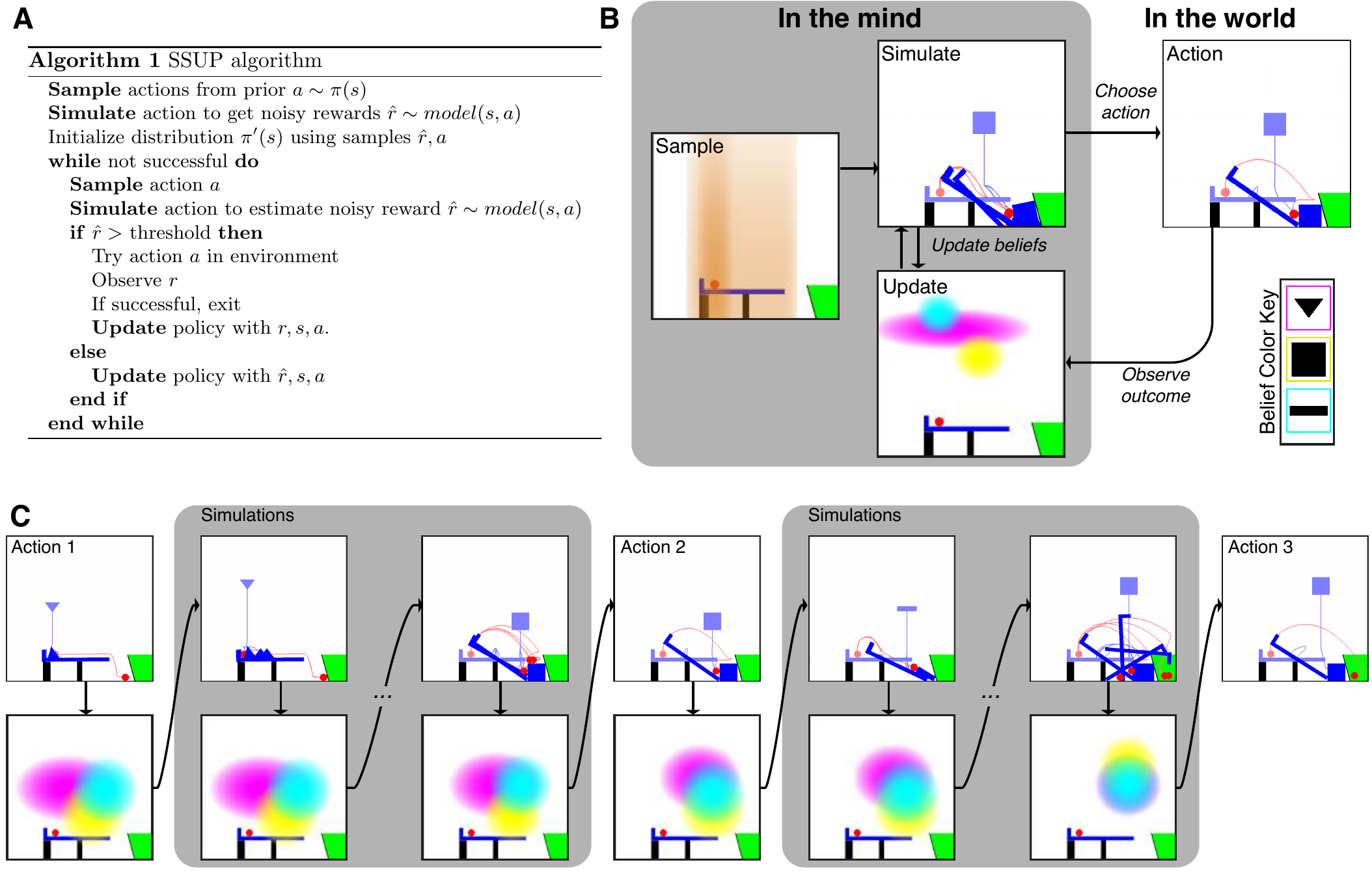}
    \caption{
    \textbf{(A)} The SSUP algorithm. 
    \textbf{(B)} A diagram of the model for the Virtual Tools game. It incorporates an object-based prior, a simulation engine for filtering proposals, and an update module that suggests new proposals based on observations ``in the mind'' and from actions taken in the world.
    \textbf{(C)} Illustration of the policy $\pi'$ evolving while attempting a level. Colored patches represent the Gaussian policy for each tool. 
    \vspace{-1.5em}
    }
    \label{fig:model}
\end{figure*}

We consider the components required to capture both the flexibility and efficiency of human tool use.
We propose that people achieve flexibility through an internal mental model that allows them to imagine the effects of actions they may have never tried before (``Simulate''). However, a mental model alone is not sufficient -- there are far too many possible actions that could be simulated, many of which are uninformative and unlikely to achieve a specific goal. Some mechanism for guiding an internal search is necessary to focus on useful parts of the hypothesis space. We therefore propose people use structured, object-oriented priors (``Sample'') and a rapid belief updating mechanism (``Update'') to guide search towards promising hypotheses. We formalize human tool use with these components as the ``Sample, Simulate, Update'' model (SSUP; Fig.~\ref{fig:model}A).

SSUP is inspired by the theory of ``problem solving as search'' \cite[][]{newell1972human}, as well as Dyna and other model-based policy optimization methods \cite[][]{sutton1991dyna, deisenroth2013survey}.
Crucially, we posit that structured priors and physical simulators must already be in place in order to solve problems as rapidly as people;
thus unlike most model-based policy optimization methods, we do not perform online updates of the dynamics model.

We want to emphasize that we view SSUP as a general modeling framework for physical problem solving, and only present here one instance of that framework: the minimal model (described below, with more detail in SI Appendix, Section S2) that we think is needed to capture basic human behavior in the Virtual Tools game. 
In the discussion we highlight ways the model will need to be improved in future work, as well as aspects of physical reasoning that rely on a richer set of cognitive systems going beyond the framework presented here. 

\paragraph{Sample: object-based prior}
At a minimum, the actions we should consider to achieve any goal should have the potential to impact our environment.
We therefore incorporate an object-based prior for sampling actions.
Specifically, the model selects one of the movable objects in the scene, then chooses an x-coordinate in an area that extends slightly beyond the width of the object, and a y-coordinate either above or below that object (Fig.~\ref{fig:model}B: Prior).
For tool choice, we assume participants are equally likely to choose any of the three tools since all tools in the game were designed to be unfamiliar to participants.
Samples from this distribution are used to initialize search.

\paragraph{Simulate: a noisy physics engine}
In order to determine which sampled actions are worth trying in the world, we assume people use an ``Intuitive Physics Engine'' \cite[][]{battaglia2013simulation} to flexibly imagine the effects of their actions. This engine is able to simulate the world forwards in time with approximately correct but stochastic dynamics \cite[][]{smith2013sources,sanborn2013reconciling}. Determining the effect of a proposed action therefore involves applying that action to one's mental representation, and using the Intuitive Physics Engine to posit the range of ways that action might cause the world to unfold \cite[][]{craik1943nature,dasgupta2018learning}.
Here we implement simulation using a game physics engine with noisy dynamics.
People characteristically have noisy predictions of how collisions will resolve \cite[][]{smith2013sources}, and so for simplicity we assume uncertainty about outcomes is driven only by noise in those collisions (the direction and amount of force that is applied between two colliding objects).\footnote{We also considered models with additional sources of physics model uncertainty added, but found that the additional parameters did not improve model fit, so we do not analyze those models here.}

Since the internal model is imperfect, to evaluate an action we produce a small number of stochastic simulations ($n_{sims}$, set here at $4$) to form a set of hypotheses about the outcome.
To formalize how good an outcome is (the \emph{reward} of a given action), we borrow an idea from the causal reasoning literature for how people conceptualize ``almost'' succeeding  \cite[][]{gerstenberg2015how}. ``Almost'' succeeding is not a function of the absolute distance an action moved you towards your goal, but instead how much of a \emph{difference} that action made.
To capture this, the minimum distance between the green goal area and any of the red goal objects is recorded; these values are averaged across the simulations and normalized by the minimum distance that would have been achieved if no tool had been added.
The reward used in SSUP is 1 minus the normalized distance, so that closer objects lead to higher reward.

Once the model finds a good enough action (formalized as the average reward being above some threshold), it takes that action ``in the world.''
Additionally, to model time limits for thinking, if the model considers more than $T$ different action proposals without acting (set here at 5), it takes the best action it has imagined so far.
We evaluate the effect of all parameter choices in a sensitivity analysis (see SI Appendix, Fig.~S1).

\paragraph{Update: learning from thoughts and actions}
So far we have described a way of intelligently initializing  search to avoid considering actions that will not be useful. But what if the prior still presents an intractably large space of possible actions?

To tackle this, we incorporate an update mechanism that learns from both simulated and real experience to guide future search towards more promising regions of the hypothesis space \citep{juechems2019where}.
This is formally defined as a Gaussian mixture model policy over the three tools and their positions, $\pi'(s)$, which represents the model's belief about high value actions for each tool.
$\pi'(s)$ is initialized with samples from the object-oriented prior, and updated using a simple policy gradient algorithm \cite[][]{williams1992simple}.
This algorithm will shape the posterior beliefs around areas to place each tool which are expected to move target objects close to the goal, and are therefore likely to contain a solution. 
Such an update strategy is useful when it finds high value actions that are nearby successful actions, but may also get stuck in local optima where a successful action does not exist.
We therefore use a standard technique from reinforcement learning: epsilon-greedy exploration.
With epsilon-greedy exploration, potential actions are sampled from the policy 1 - $\epsilon$\% of the time, and from the prior $\epsilon$\% of the time.
Note that this exploration is only used for proposing internal simulations; model actions are chosen based on the set of prior simulation outcomes.
This is akin to thinking of something new, instead of focusing on an existing strategy.

\section*{Results}
We analyze human performance on the first 20 levels of the Virtual Tools game and compare humans to the SSUP model and alternates, including SSUP models with ablations and two alternate learning baselines. We show that the full SSUP model best captures human performance.
Access to the game and all data including human and model placements is provided at \url{https://sites.google.com/view/virtualtoolsgame}.
\begin{figure*}[t!]
\centering
    \includegraphics[width=0.9\linewidth]{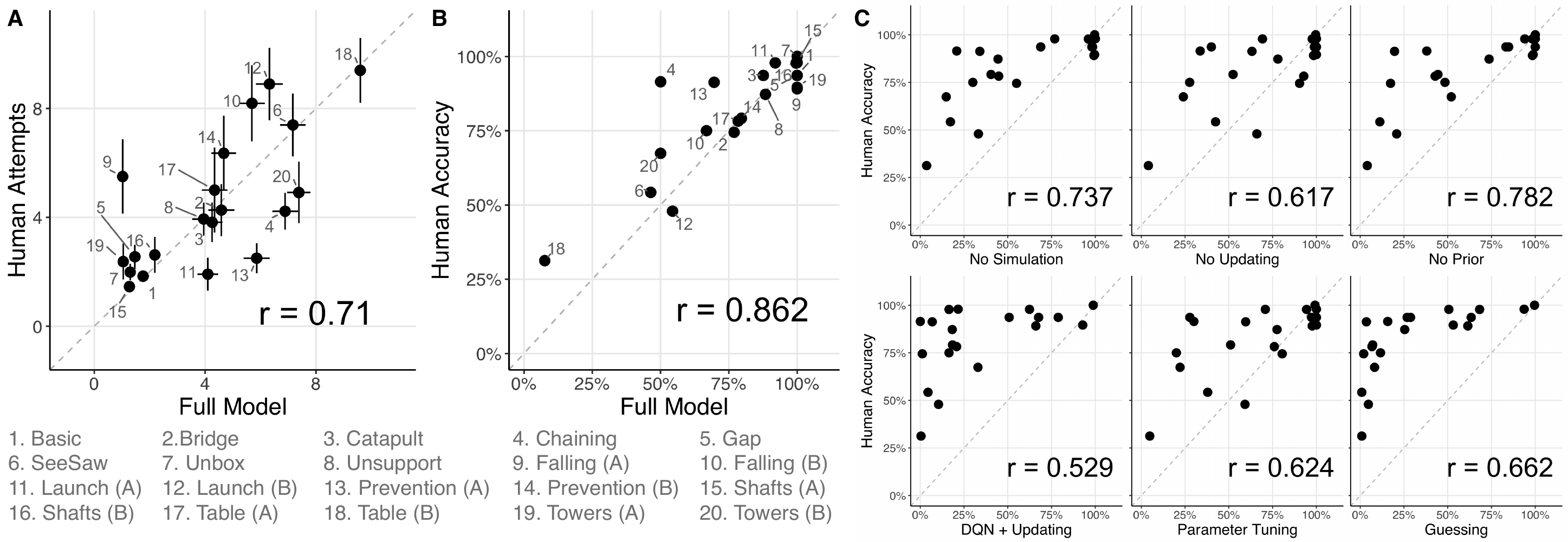}
    \caption{\textbf{(A)} Comparison of average number of human participants' attempts for each level with average number of attempts for the SSUP model. Bars indicate $95\%$ confidence intervals on estimates of the means.
    \textbf{(B)} Comparison of human participants' accuracy on each trial versus the accuracy of the SSUP model. 
    \textbf{(C)} Comparison of human participants' accuracy to all alternate models. Numbers correspond to the trials in Fig.~\ref{fig:trials}.}
    \vspace{-1em}
    \label{fig:abl_compare}
\end{figure*}

\subsection*{Human results} \label{sec:empresults}
Experiments were approved by the MIT Committee on the Use of Humans as Experimental Subjects under protocol \#0812003014. Participants were notified of their rights before the experiment, were free to terminate participation at any time by closing the browser window, and were compensated monetarily for their time.

We recruited 94 participants through Amazon Mechanical Turk and asked each participant to solve 14 levels: all 8 of the unmatched levels, and one variation of each of the 6 matched pairs (randomly selected). 

Participants could choose to move on once a problem was solved, or after two minutes had passed.
See SI Appendix, Section S1 for further details.

The variation in difficulty between levels of the game was substantial. 
Participants showed an average solution rate of 81$\%$ (sd = 19\%), with the range covering 31$\%$ for the hardest level to 100$\%$ for the easiest.
Similarly, participants took an average of 4.5 actions (sd = 2.5) for each level, with a range from 1.5 to 9.4 average attempts.
Even within trials, there was a large amount of heterogeneity in the number of actions participants used to solve the level. This would be expected with ``rapid trial-and-error'' learning: participants who initially tried a promising action would solve the puzzle quickly, while others explored different actions before happening on promising ones (e.g., Fig.~\ref{fig:slowfast}).

Behavior differed across all six matched level pairs.
We study whether these subtle differences do indeed affect behavior, even without feedback on the first action, by asking whether we can identify which level variant each action came from.
We find these actions are differentiable across matched levels in `Shafts', `Prevention', `Launch' and `Table' on the first attempt, but not `Falling' or `Towers' (see SI Appendix, Fig. S11 and Sec. S6A for details).
However, participants required a different number of actions to solve every level (all $ts > 2.7$, $ps < 0.01$).
This suggests that people are paying attention to subtle differences in the scene or goal to choose their actions.

\subsection*{Model results}

We investigate several metrics for comparing the models to human data.
First, we look at how quickly and how often each model solves each level, and whether that matches participants.
This is measured as the correlation and root mean squared error (RMSE) between the average number of participant attempts for each level and the average number of model attempts for each level, and the correlation and RMSE between human and model solution rates. 
The SSUP model explains the patterns of human behavior across the different levels well (SI Appendix, Table~S2). It uses a similar number of attempts on each level ($r = 0.71$; $95\% ~\text{CI} = [0.62, 0.76]$; mean empirical attempts across all levels: $4.48$, mean model attempts: $4.24$; Fig.~\ref{fig:abl_compare}A) and achieves similar accuracy ($r = 0.86$; $95\%~\text{CI} = [0.76, 0.89]$; Fig.~\ref{fig:abl_compare}B).

\begin{figure*}[h!]
    \centering
    \includegraphics[width=.9\textwidth]{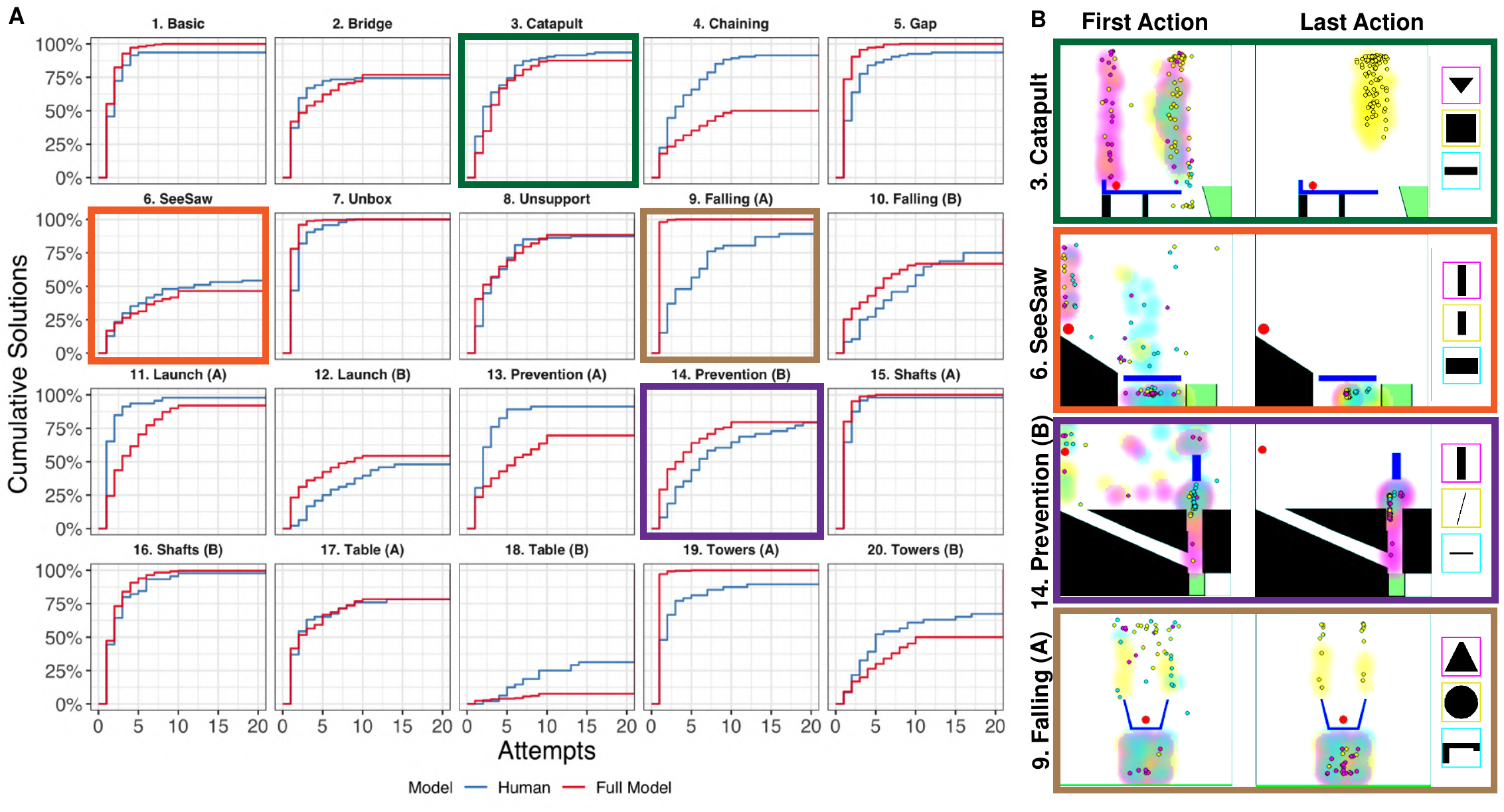}
    \caption{\textbf{(A)} Cumulative solution rate over number of placements for participants vs. the SSUP model. \textbf{(B)} Distribution of model actions (background) versus human actions (points) on the first and last attempts of the level for a selection of four levels. The distribution of model actions is estimated based on fitting a Kernel Density Estimate to the actions taken by the model across 250 simulations. Colors indicate the tool used, with the tools and associated colors shown to the right of each level. In most levels, the SSUP model captures the evolution of participants' solutions well, including the particular actions chosen; in the few cases that it differs, there is no alternative model that systematically explains these differences.
    }
    \vspace{-1em}
    \label{fig:abcs_compare}
\end{figure*}

Across many levels, the SSUP model not only achieves the same overall solution rate as people, but approaches it at the same rate.
We measure this by looking at cumulative solution rates -- over all participants or model runs, what proportion solved each level within $X$ placements -- and find that people and the model often demonstrate similar solution profiles (Fig.~\ref{fig:abcs_compare}A; see SI Appendix, Section S6B for quantitative comparison).

We can look in more detail how the model accomplishes this by comparing both the first actions that people and the model takes (Fig.~\ref{fig:abcs_compare}B), and the actions that both take to solve a level (Fig.~\ref{fig:abcs_compare}C).
Like our human participants, the model takes significantly different actions on the first attempt between matched level pairs (see SI Appendix, Sec. S6A).
More generally, both people and the model will often begin with a variety of plausible actions (e.g., Catapult). 
In some cases, both will attempt initial actions that have very little impact on the scene (e.g., SeeSaw and Prevention (B)); this could be because people cannot think of any useful actions and so decide to try \emph{something}, similar to how the model can exceed its simulation threshold. However, in other cases, the model's initial predictions diverge from people, and this leads to a different pattern of search and solutions. For instance, in Falling (A), the model quickly finds that placing an object under the container will reliably tip the ball onto the ground, but people are biased to drop an object from above. Because of this, the model often rapidly solves the level with an object below, whereas a proportion of participants find a way to flip the container from above; this discrepancy can also be seen in the comparison of number of attempts before the solution, where the model finds a solution quickly, while people take a good deal longer (Fig.~\ref{fig:abl_compare}A). 
For comparisons of the first and last actions across all levels, see SI Appendix, Fig. S11.

\subsubsection*{Model comparisons on Virtual Tools} \label{sec:modcompare}

We compare the full SSUP model against a set of six alternate models. Three models investigate the contribution of each SSUP component by removing the prior, simulation, or updating individually. Two models propose alternate solution methods: learning better world models rather than learning over actions (Parameter Tuning) or replacing the prior and simulator with a learned proposal mechanism (DQN + Updating). 
The Parameter Tuning alternate model uses inference to learn object densities, frictions and elasticities from observed trajectories. 
The learned proposal mechanism corresponds to a model-free deep reinforcement learning agent \cite[][]{mnih2015human} which is trained on a set of 4500 randomly generated levels of the game (see SI Appendix, Sec.~S5), and then updated online for each of the 20 testing levels using the same mechanism as SSUP. 
This model has substantially more experience with the environment than other models, and serves as a test of whether model-free methods can make use of this experience to learn generalizable policies that can guide rapid learning.
Finally, we compare to a ``Guessing'' baseline for performance if an agent were to simply place tools randomly. See Fig.~\ref{fig:abl_compare}C and SI Appendix, Table S2 for these comparisons.

Eliminating any of the three SSUP components causes a significant decrease in performance (measured as deviation between empirical and model cumulative solution curves; all bootstrapped $ps<0.0001$; see SI Appendix, Sec. S6B, Fig. S6 for further detail). The reduced models typically require more attempts to solve levels because they are either searching in the wrong area of the action space (No Prior), attempting actions that have no chance of being successful (No Simulation), or do not guide search towards more promising areas (No Updating).

DQN + Updating performs worst of all plausible alternate models, using the most actions and solving levels at a rate barely over chance.
Because this is equivalent to the No Simulation model with a different prior, its poor performance suggests that generalized action policies cannot easily be learned from repeatedly playing similar levels (see SI Appendix, Sec.~S5).

Because the Parameter Tuning model is equivalent to the No Updating model except that the properties of the dynamics model can be learned in Parameter Tuning, comparing those two models allows us to test whether we need to assume that people are learning the dynamics of the world in this game.
The fact that both models perform roughly equivalently (see Fig.~\ref{fig:abl_compare}C) suggests that we do not need this assumption here.

\begin{figure*}[h!]
    \centering
    \includegraphics[width=.9\textwidth]{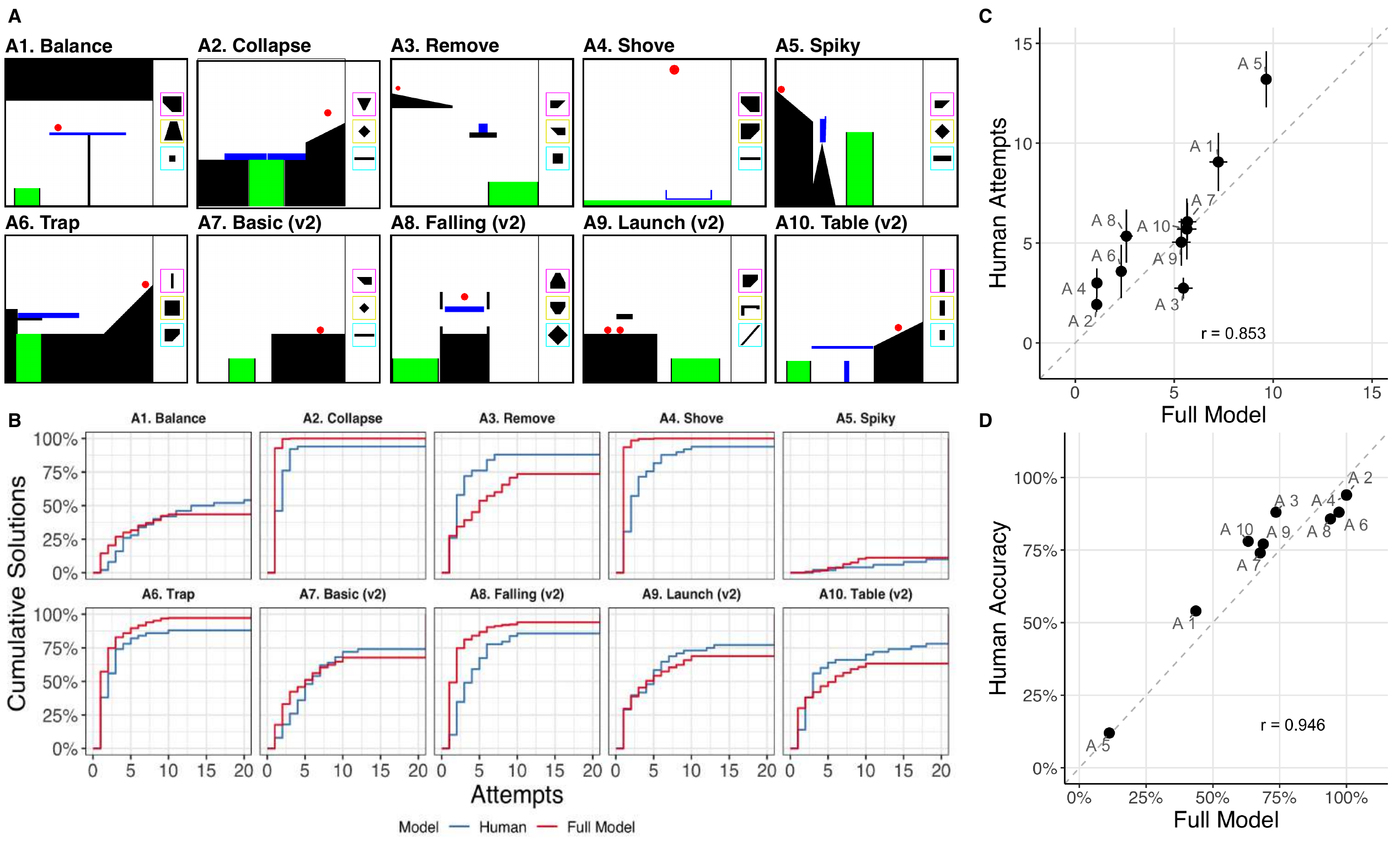}
    \vspace{0.1em}
    \caption{Results on 10 additional trials. \textbf{(A)} Trials used for the second experiment. \textbf{(B)} The cumulative solution rate for participants and the SSUP model. 
    \textbf{(C)} Comparison of the number of human and model actions by trial.
    \textbf{(D)} Comparison of human and model accuracy on each trial.
    }
    \vspace{-1em}
    \label{fig:cv_fig}
\end{figure*}

Finally, we quantified how well each model captured the particular actions people took. Due to heterogeneity in participants' responses, we were unable to cleanly differentiate models' performance except to find that the DQN + Updating model underperformed the rest (see SI Appendix, Sec. S6C). However, no model reached the theoretical noise ceiling, suggesting components of the SSUP framework could be improved to better explain participants' actions (see the Discussion).

\subsection*{Validation on novel levels}
We conducted a second experiment to test whether the models generalize to novel levels and physical concepts without tuning hyperparameters. For this experiment, we created 10 new levels: 6 novel level types and 4 variants of the originals (Fig \ref{fig:cv_fig}A), testing an independent sample of $50$ participants on all levels. The 6 novel level types were designed to test new physical strategies, including balancing, breaking, and removing objects from a ball's path. All other experimental details were identical to the main experiment.

Without tuning any model parameters, we find a good correspondence between human and model solution rates (Fig.~\ref{fig:cv_fig}B), and a strong correlation between the model's performance and human performance across number of placements (Fig.~\ref{fig:cv_fig}C, $r=0.85$) and accuracy (Fig.~\ref{fig:cv_fig}D, $r = 0.95$).
Similar to the main experiment, we find a decrement in performance if the prior or simulation are removed, or for the DQN + Updating model (all bootstrapped $ps<0.0001$; SI Appendix, Fig. S7). However, while numerically worse, we do not find a reliable difference if the update mechanism is removed ($p=0.055$) or swapped for model learning ($p=0.346$), suggesting that the particular reward function or update procedure might be less applicable to these levels (see SI Appendix, Sec. S6B).

\section*{Discussion}
We introduce the Virtual Tools game for investigating flexible physical problem solving in humans and machines, and show that human behavior on this challenge expresses a wide variety of trial-and-error problem solving strategies.
We also introduce a model for human physical problem solving: ``Sample, Simulate, Update.'' The model presumes that to solve these physics problems, people rely on an internal model of how the world works. Learning in this game therefore involves condensing this vast world knowledge to rapidly learn how to act in each instance, using a structured trial-and-error search. 

\subsection*{Model limitations}

Although the SSUP model we used solves many of the levels of the Virtual Tools game in a human-like way, we believe that this is still only a first approximation to the rich set of cognitive processes that people bring to the task.  In particular, there are at least two ways in which the model is insufficient: its reliance on very simple priors, and its planning and generalizing only in the forwards direction.

\begin{figure}[t]
    \centering
    \includegraphics[width=0.8\columnwidth]{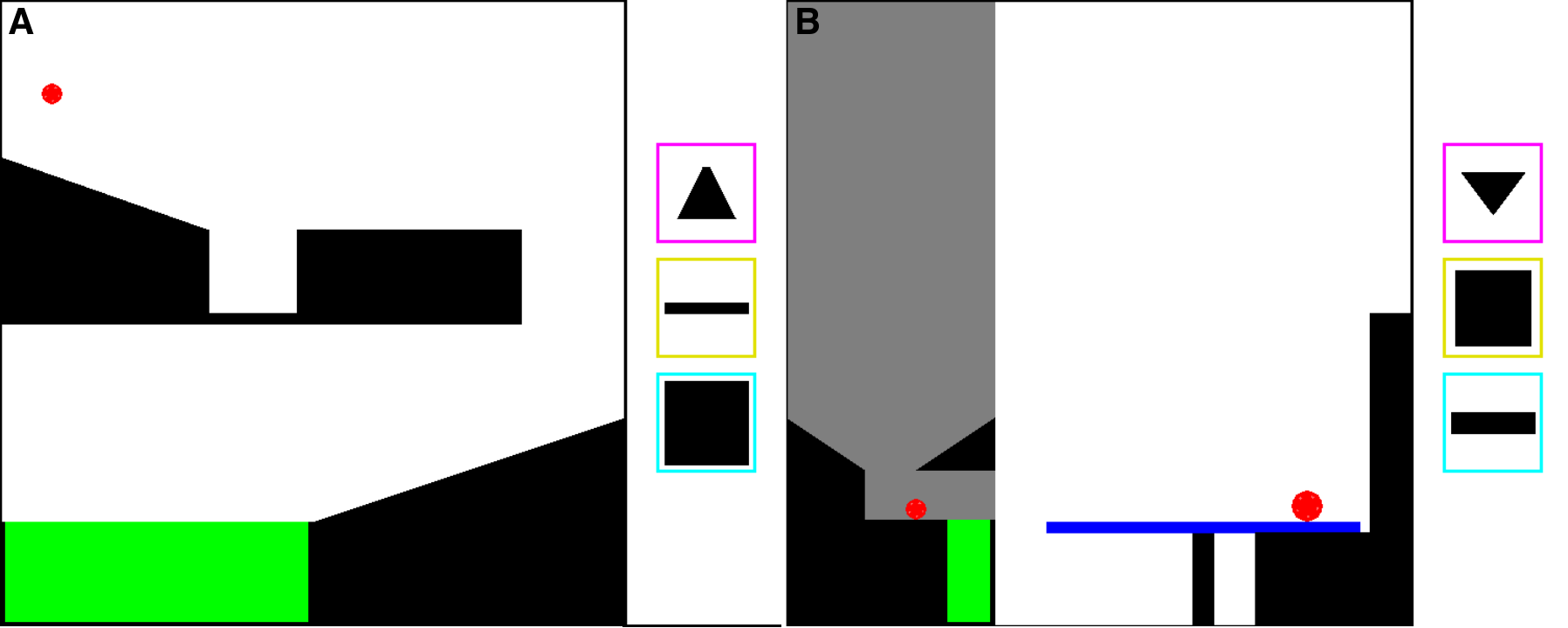}
    \caption{Two problems that demonstrate limitations of the current model. \textbf{(A)} A ``suspicious conicidence'' that one tool fits perfectly in the hole. \textbf{(B)} Creating a `sub-goal' to launch the ball onto the other side is useful.}
    \vspace{-1em}
    \label{fig:limits}
\end{figure}

We can see the limits of the object-based prior in the Falling (A) level (Fig.~\ref{fig:abl_compare}B): people are much less likely to consider placing an object underneath the container to tip it over. Instead, many people try to tip it over from above, even though this is more difficult. In this way, people's priors over strategies are \emph{context specific}, which causes them to be slower than the model in this level.
In other cases, this context specificity is helpful: for instance, in the hypothetical level shown in Fig.~\ref{fig:limits}A, there is a hole that one of the tools fits suspiciously perfectly into. Many people notice this coincidence quickly, but because the model cannot assess how tools might fit into the environment without running a simulation, it only succeeds 10\% of the time. In future work, a more complex prior could be instantiated in the SSUP framework, but it remains an open question how people might form these context-specific priors, or how they might be shaped over time via experience.

People show greater flexibility than our model in the ability to work backwards from the goal state to find more easily solvable sub-goals \cite[][]{anderson1993problem}. 
In the hypothetical level in Fig.~\ref{fig:limits}B, the catapult is finicky, which means that most catapulting actions will not make it over the barrier, and therefore will never hit the ball on the left. Instead, the easiest way to increase the objective function is by the incorrect strategy of knocking the ball on the right to get close to the goal, and therefore the model only solves the level 8\% of the time. Working backwards to set the first sub-goal of launching the ball over the barrier would prevent getting stuck with knocking the ball as a local minimum. From an engineering standpoint, creating sub-goals is natural with discrete problem spaces \cite[][]{newell1972human}, but it is less clear how these might be discovered in the continuous action space of the Virtual Tools game.

\subsection*{Related cognitive systems}

There is an extensive body of research into the cognitive systems that underlie the use of real-world tools, including understanding how to manipulate them and knowing their typical uses \cite[e.g.][]{vaesen2012cognitive,orban2014neural,osiurak2016tool,beck2011making}.
Here our focus was on ``mechanical knowledge'' of tools: how to use objects in novel situations. However, in real-world tool use, these systems work together with motor planning and semantic knowledge of tools. Future work can focus on these links, such as how novel tools become familiar, or how our motor limits constrain the plans we might consider.

The Virtual Tools game presents a problem solving task that blends facets of prior work, but encompasses a novel challenge.
To rapidly solve these problems requires good prior knowledge of the dynamics -- unlike Complex Problem Solving in which the dynamics are learned in an evolving situation \cite[][]{frensch1995complex} -- and further iteration once a promising solution is considered -- unlike the `a-ha' moment that leads immediately to a solution in Insight Problem Solving \cite[][]{chu2011human, gick1980analogical}.
Unlike in traditional model-based or model-free reinforcement learning, in this task people bring rich models of the world that they can quickly tailor to specific, novel problems.

Distilling rich world knowledge to useful task knowledge is necessary for any agent interacting with a complex world.
One proposal for how this occurs is ``learning by thinking'' \cite[][]{lombrozo2018learning}: translating knowledge from one source (internal models of physics) to another, more specific instantiation (a mapping between actions and outcomes on \emph{this particular} level).
We show how SSUP instantiates one example of ``learning by thinking'': by training a policy with data from an internal model.
Evidence for this sort of knowledge transfer has been found in people \cite[][]{gershman2014retrospective,gershman2017imaginative}, but has focused on simpler discrete settings in which the model and policy are jointly learned.

\subsection*{Virtual Tools as an AI Challenge}
In preliminary experiments with model-free reinforcement learning approaches \cite[][]{mnih2015human}, we found limited generalization with inefficient learning across almost all of the Virtual Tools levels (see SI Appendix, Section S5) despite significant experience with related levels. 

Based on our human experiments, we believe that model-based approaches will be required to be able to play games like Virtual Tools.
Such approaches are becoming increasingly popular in machine learning \cite[][]{weber2017imagination}, especially when combined with ``learning-to-learn'' techniques that can learn to adapt quickly to new tasks \cite[][]{finn2017model,schmidhuber1998reinforcement}.
Learning these models remains challenging, but approaches that incorporate added structure have excelled in recent years \cite[][]{chang2016compositional,jayaraman2018time}.
Within the AI and robotics communities, model-based methods are already popular \cite[][]{toussaint2018differentiable,kaelbling2010hierarchical,garcia1989model}. 
Remaining challenges include how to learn accurate enough models that can be used with raw sensor data \cite[][]{kroemer2019review}, and how to handle dynamic environments.

Virtual Tools adds to a growing set of environments that test artificial agents' abilities to predict and reason using physics, such as the concurrently developed PHYRE benchmark \cite[][]{bakhtin2019phyre} and others \cite[][]{wenke2019reasoning,bapst2019structured,ge2016hole}. In contrast, our focus is on providing problems that people find challenging but intuitive, where solutions are non-obvious and do not rely on precise knowledge of world dynamics. By contributing human data to compare artificial and biological intelligence, we hope to provide a test-bed for more human-like artificial agents.

\subsection*{Future empirical directions}

This work provides an initial foray into formalizing the computational and empirical underpinnings of flexible tool use, but there remains much to study.
For instance, we do not find evidence that people learn more about the world, perhaps because here there is little benefit to additional precision here. But there are cases where learning the dynamics is clearly helpful (e.g., discovering that an object is abnormally heavy, or glued down), and we \emph{would} expect people to update their physical beliefs in these cases. When and in what ways people update their internal models to support planning is an important area of study.

Children can discover how to use existing objects earlier than they can make novel tools \cite[][]{beck2011making}, suggesting that tool creation is more challenging than tool use. Yet it is the ability to make and then pass on novel tools that is theorized to drive human culture \cite[][]{tomasello1999cultural}. It is therefore important to understand not just how people use tools, but also how they develop and transmit them, which we can study by expanding the action space of the Virtual Tools game.

\subsection*{Conclusion}
Understanding how to flexibly use tools to accomplish our goals is a basic and central cognitive capability. In the Virtual Tools game, we find that people efficiently use tools to solve a wide variety of physical problems. 
We can explain this rapid trial-and-error learning with the three components of the SSUP framework: rich prior world knowledge, simulation of hypothetical actions, and the ability to learn from both simulations and observed actions.
We hope this empirical domain and modeling framework can provide the foundations for future research on this quintessentially human trait: using, making, and reasoning about tools, and more generally shaping the physical world to our ends.

\subsection*{Acknowledgments}

The authors would like to thank Leslie Kaelbling, Roger Levy, Eric Schulz, Jessica Hamrick, and Tom Silver for helpful comments, and Mario Belledonne for help with the Parameter Tuning model. This work was supported by NSF STC award CCF-1231216, ONR MURI N00014-13-1-0333, and research grants from ONR, Honda, and Mitsubishi Electric.

\bibliography{biblio}

\end{document}